\documentclass[10pt,twocolumn,letterpaper]{article}
\usepackage{iccv}
\usepackage{times}
\usepackage{epsfig}
\usepackage{graphicx}
\usepackage{amsmath}
\usepackage{amssymb}
\usepackage{amsfonts}
\usepackage{mathtools}
\usepackage[accsupp]{axessibility}

\usepackage{bm}

\usepackage{booktabs}
\usepackage{enumitem}
\usepackage[ruled, vlined, linesnumbered]{algorithm2e}
\usepackage{xscommand}
\usepackage{subcaption}

\usepackage{gensymb}
\usepackage[pagebackref,breaklinks,colorlinks]{hyperref}
\usepackage{caption}
\usepackage{subfloat}

\iccvfinalcopy 

\def\iccvPaperID{****} 

\ificcvfinal\fi

\begin{document}

\title{Structure Invariant Transformation for better Adversarial Transferability}

\author{Xiaosen Wang\\
Huawei Singularity Security Lab\\
{\tt\small xiaosen@hust.edu.cn}
\and
Zeliang Zhang\\
HUST\\
{\tt\small hust0426@gmail.com}
\and
Jianping Zhang\\
Chinese University of Hong Kong\\
{\tt\small jpzhang@cse.cuhk.edu.hk}
}

\maketitle
\ificcvfinal\thispagestyle{empty}\fi

\begin{abstract}
      Given the severe vulnerability of Deep Neural Networks (DNNs) against adversarial examples, there is an urgent need for an effective adversarial attack to identify 
      the deficiencies of DNNs in security-sensitive applications. As one of the prevalent black-box adversarial attacks, the existing transfer-based attacks still cannot achieve comparable performance with the white-box attacks. Among these, input transformation based attacks have shown remarkable effectiveness in boosting transferability. In this work, we find that the existing input transformation based attacks transform the input image globally, resulting in limited diversity of the transformed images. We postulate that the more diverse transformed images result in better transferability. Thus, we investigate how to locally apply various transformations onto the input image to improve such diversity while preserving the structure of image. To this end, we propose a novel input transformation based attack, called Structure Invariant Attack (\name), which applies a random image transformation onto each image block to craft a set of diverse images for gradient calculation. Extensive experiments on the standard ImageNet dataset demonstrate that \name exhibits much better transferability than the existing SOTA input transformation based attacks on CNN-based and transformer-based models, showing its generality and superiority in boosting transferability. Code is available at \url{https://github.com/xiaosen-wang/SIT}.
\end{abstract}

\section{Introduction}
\label{sec:intro}
With the unprecedented progress of Deep Neural Networks (DNNs)~\cite{alex2012ImageNet,he2016resnet,huang2017densely,vaswani2017attention,dosovitskiy2020image},
they have been deployed in many security-sensitive applications, such as face recognition~\cite{florian2015facenet,wang2018cosface,song2022adaptive,song2022face}, autonomous driving~\cite{geiger2012we,lillicrap2015continuous}, \etc. On the other hand, recent works have found that DNNs are vulnerable to adversarial examples~\cite{szegedy2014intriguing,goodfellow2015FGSM}, which mislead the deep models with imperceptible perturbations. This brings a huge threat to the real-world applications~\cite{sharif2016accessorize,eykholt2018robust,song2021tacr, wang2022triangle,yuan2022natural,zhang2022practical} and makes it imperative for an effective attack to identify the deficiencies of DNNs when we robustify the deep models or deploy them for commercial applications.

Existing adversarial attacks usually fall into two categories: \textit{white-box attacks}~\cite{goodfellow2015FGSM,moosavi2016deepfool,madry2018pgd,kurakin2017adversarial,wang2019gan} can fetch any information of the target model, including (hyper-)parameters, gradient, architecture, while \textit{black-box attacks}~\cite{ilyas2018black,cheng2019improving,brendel2018decision,li2020qeba,liu2016delving,zhang2022beyond} are only allowed limited access to the target model. One of the significant properties of adversarial examples is their transferability~\cite{xie2019improving,dong2018boosting,wang2021enhancing,zhang2023improving}, in which the adversarial examples generated on one model can still mislead other models, making it possible to attack the real-world applications in the black-box setting. However, existing adversarial attacks~\cite{kurakin2017adversarial,madry2018pgd} often exhibit superior white-box attack performance but poor transferability.

\begin{figure}
    \centering
    \begin{minipage}[c]{0.10\textwidth} 
        \begin{subfigure}{\textwidth}
          \centering 
          \includegraphics[width=\linewidth]{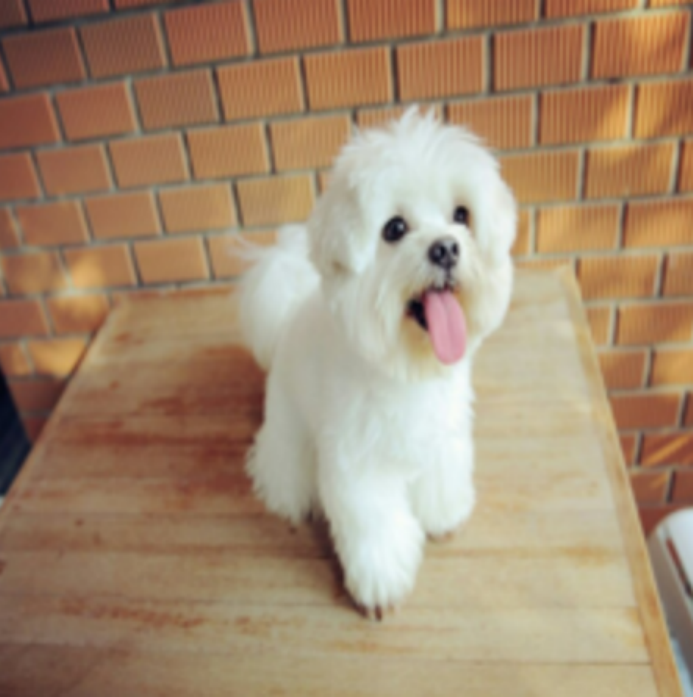}
          \vspace{-1.3em}
          \caption*{Raw Image}
        \end{subfigure}\\
    \end{minipage}
    \hspace{0.1cm}
    \begin{minipage}[c]{0.10\textwidth} 
        \begin{subfigure}{\textwidth}
          \centering 
          \includegraphics[width=\linewidth]{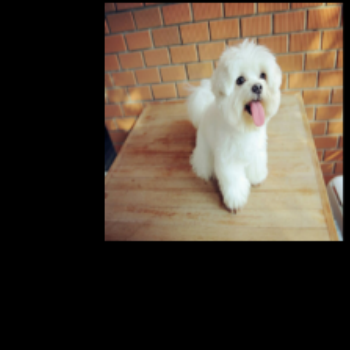}
          \vspace{-1.3em}
          \caption*{DIM}
        \end{subfigure}\\
        \begin{subfigure}{\textwidth} 
          \centering 
          \includegraphics[width=\linewidth]{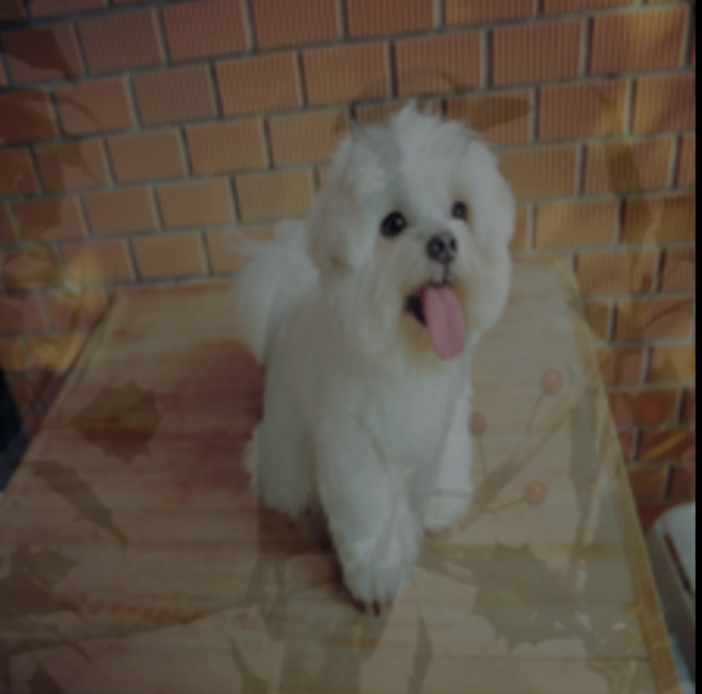}
          \vspace{-1.3em}
          \caption*{Admix}
        \end{subfigure}%
    \end{minipage}
    \hspace{0.1cm}
    \begin{minipage}[c]{0.10\textwidth} 
        \begin{subfigure}{\textwidth}
          \centering 
          \includegraphics[width=\linewidth]{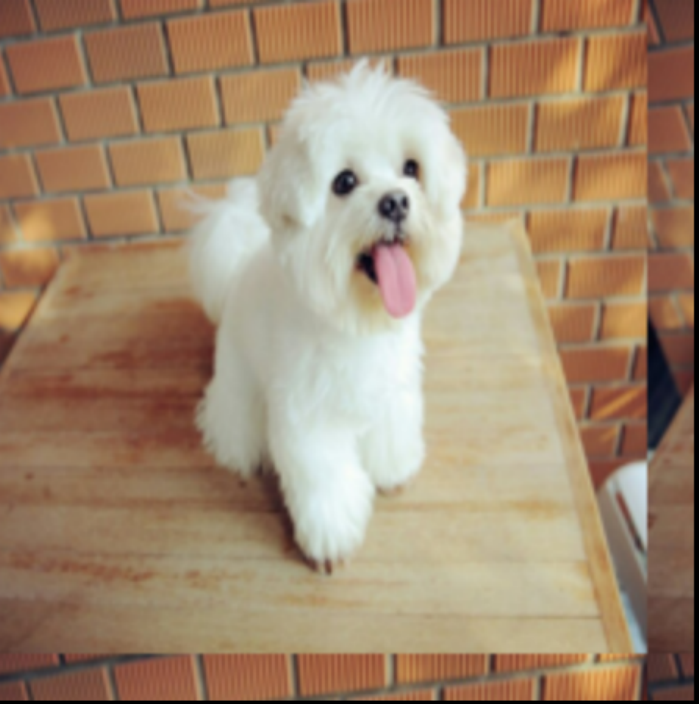}
          \vspace{-1.3em}
          \caption*{TIM}
        \end{subfigure}\\
        \begin{subfigure}{\textwidth} 
          \centering 
          \includegraphics[width=\linewidth]{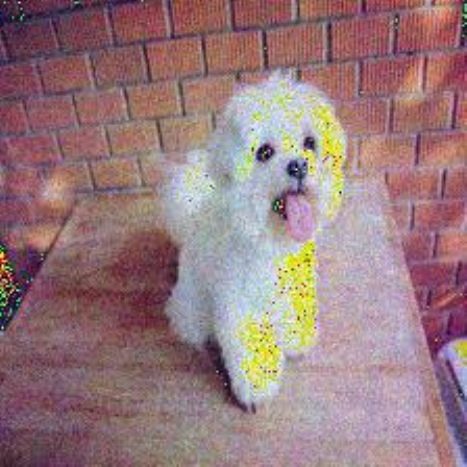}
          \vspace{-1.3em}
          \caption*{SSA}
        \end{subfigure}%
    \end{minipage}
     \hspace{0.1cm}
    \begin{minipage}[c]{0.10\textwidth} 
        \begin{subfigure}{\textwidth}
          \centering 
          \includegraphics[width=\linewidth]{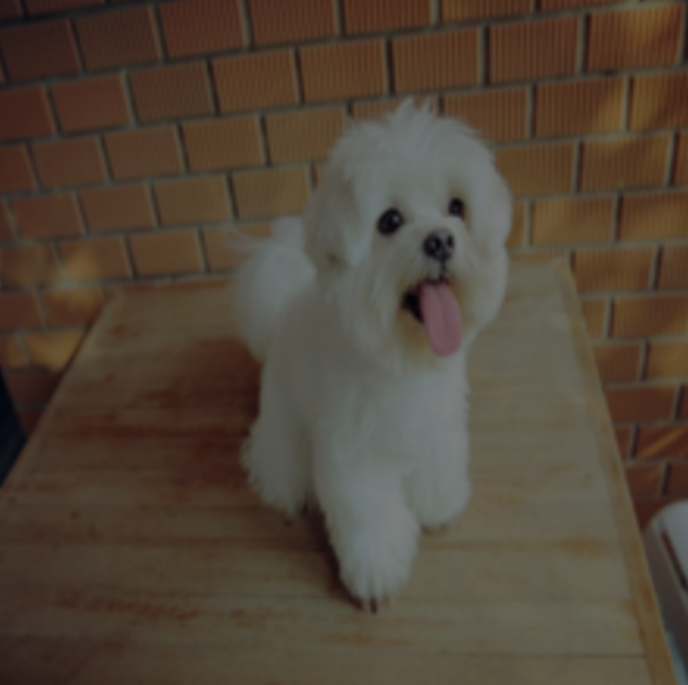}
          \vspace{-1.3em}
          \caption*{SIM}
        \end{subfigure}\\
        \begin{subfigure}{\textwidth} 
          \centering 
          \includegraphics[width=\linewidth]{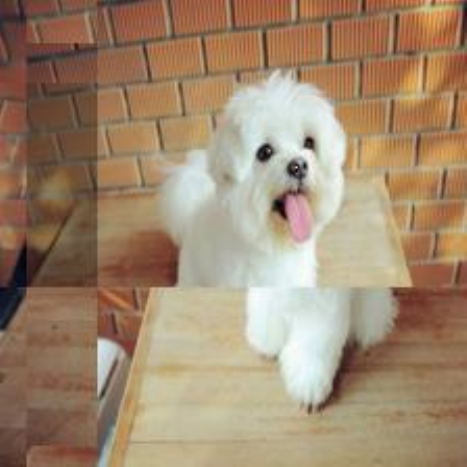}
          \vspace{-1.3em}
          \caption*{\name (Ours)}
        \end{subfigure}%
    \end{minipage}
    \caption{The raw image and its transformed images by DIM (resize factor of $0.8$)~\cite{xie2019improving}, TIM (translated by $15$ pixels)~\cite{dong2019evading}, SIM (scale factor of $0.5$), \textit{Admix} (admix strength of $0.2$ and scale factor of $0.5$)~\cite{wang2021admix}, SSA (turning factor of $0.5$)~\cite{long2022frequency} and our proposed \name ($3\times3$ blocks).}
    \label{fig:transformation}
\end{figure} 

To craft more transferable adversarial examples, various techniques have been proposed, such as momentum-based methods~\cite{dong2018boosting,lin2020nesterov,wang2021enhancing,wang2021boosting}, input transformations~\cite{xie2019improving,dong2019evading,lin2020nesterov,yang2023adversarial,yuan2022adaptive}, ensemble attacks~\cite{liu2016delving,xiong2022stochastic,long2022frequency}, advanced objective functions~\cite{wu2020boosting,zhou2018transferable} and model-specific approaches~\cite{li2020learning,wu2020skip}. Among which, input transformations (\eg, random resizing and padding~\cite{xie2019improving}, translation~\cite{dong2019evading}, scale~\cite{lin2020nesterov}
admix~\cite{wang2021admix}, \etc) that transform the image before gradient calculation, have achieved superior transferability and attracted broad attention. Nevertheless, we find that the existing input transformation based attacks, even the SOTA attack \textit{Admix}~\cite{wang2021admix} and SSA~\cite{long2022frequency}, transform the image globally without changing the local relationship among objects in the input image. We argue that the diversity of such transformed images is still not enough, leading to limited transferability.

In this work, we postulate and empirically validate that the more diverse transformed images lead to better transferability. Based on this observation, instead of applying a single transformation on the input image, we apply different transformations locally on different parts of the image to enhance the diversity of transformed images. As shown in Fig.~\ref{fig:transformation}, such transformation can bring much more difference to the generated image compared with the raw image but still preserve the global structure of the object. Based on this transformation, we propose a novel input transformation based attack called structure invariant attack (\name), which utilizes the gradient of these transformed images to update the adversarial examples for better transferability.

Our contributions are summarized as follows:
\begin{itemize}[leftmargin=*,noitemsep,topsep=2pt]
    \item We empirically validate that the high diversity of transformed images is beneficial to improve transferability, which sheds new light on how to design new input transformations for more transferable adversarial examples.
    \item We design a new image transformation method, which locally applies different transformations on different parts of the image to generate more diverse images but preserve its global structure.
    \item Based on the proposed image transformation, we devise a new input transformation based attack, called structure invariant attack (\name), to generate more transferable adversarial examples. 
    \item Extensive experiments on ImageNet dataset demonstrate that \name outperforms the baselines with a clear margin on CNN-based as well as transformer-based models, showing its superiority and generality.
\end{itemize}

\section{Related Work}
Adversarial examples~\cite{szegedy2014intriguing} have brought an impressive threat to the DNN-enabled applications, such as computer vision~\cite{goodfellow2015FGSM,wei2019transferable,athalye2018obfuscated}, natural language processing~\cite{alzantot2018generating,wang2021natural,yang2022robust}, speech recognition~\cite{carlini2018audio,yang2019characterizing}, \etc. To identify the vulnerability of DNNs, various attack methods have been proposed recently, such as gradient-based attacks~\cite{goodfellow2015FGSM,moosavi2016deepfool,kurakin2017adversarial,madry2018pgd}, score-based attacks~\cite{ilyas2018black,uesato2018adversarial,guo2019simple,li2019nattack,andriushchenko2020square}, decision-based attacks~\cite{brendel2018decision,cheng2019query,li2020qeba,wang2022triangle} and transfer-based attacks~\cite{liu2016delving,dong2018boosting,xie2019improving,lin2020nesterov,wang2021admix}. Among these, transfer-based attacks do not access any information of the target model, making it applicable to attack any model in the physical world. In this work, we focus on generating more transferable adversarial examples and briefly introduce the existing transfer-based attacks, the corresponding defense methods, and data augmentations.

\subsection{Adversarial Attack}
As the first gradient-based attack, Fast Gradient Sign Method (FGSM)~\cite{goodfellow2015FGSM} adds the perturbation in the gradient direction to the benign sample, leading to high attack efficiency but limited performance. Later, I-FGSM~\cite{kurakin2017adversarial} extends FGSM into an iterative version, which achieves much better white-box attack performance but poor transferability. Given the high efficiency and effectiveness of I-FGSM, numerous transfer-based attacks are proposed to boost transferability based on I-FGSM to attack the deep model in the black-box setting.

\textbf{Momentum-based methods.} MI-FGSM~\cite{dong2018boosting} introduces momentum into I-FGSM to stabilize the optimization direction and escape local maxima. NI-FGSM~\cite{lin2020nesterov} adopts Nesterov Accelerated Gradient to accumulate the momentum, which achieves better transferability. Variance tuning~\cite{wang2021enhancing} adopts the gradient variance of the previous iteration to tune the current gradient in MI-FGSM and NI-FGSM, significantly improving transferability. EMI-FGSM~\cite{wang2021boosting} enhances the momentum by accumulating several samples' gradients in the previous gradient's direction to further stabilize the optimization direction.

\textbf{Input transformation based attacks.} DIM~\cite{xie2019improving} is the first input transformation based attack, which adds padding to a randomly resized image for fixed size before gradient calculation. TIM~\cite{dong2019evading} optimizes the perturbation over an ensemble of translated images, which is further approximated by convolving the gradient at the untranslated image with a pre-defined kernel. SIM~\cite{lin2020nesterov} scales the images with different scale factors for gradient calculation. DEM~\cite{zou2020improving} averages the gradient on several diverse transformed images similar to DIM with various resizing factors. \textit{Admix}~\cite{wang2021admix} calculates the gradient on the input image admixed with a small portion of each add-in image from other categories while using the original label of the input. Wu~\etal~\cite{wu2021improving} train an adversarial transformation network to destroy the adversarial perturbation and require the synthesized adversarial examples resistant to such transformations. SSA~\cite{long2022frequency} adds Gaussian noise and randomly masks the image in the frequency domain to transform the input image.

\textbf{Ensemble attacks.} Liu~\etal~\cite{liu2016delving} first found that the adversarial examples generated on multiple models denoted as ensemble attack, are more transferable. Recently, Xiong~\etal~\cite{xiong2022stochastic} reduce the gradient variance between various models to boost the ensemble attack.

\textbf{Advanced objective functions.} The above attacks often take the cross-entropy loss as the objective function. Researchers also find that some regularizers are beneficial to boost transferability. For instance,  Zhou~\etal~\cite{zhou2018transferable} additionally maximize the difference of the intermediate feature maps between the benign sample and adversarial example. Wu~\etal~\cite{wu2020boosting} adopt a regularizer about the distance of the attention maps between these samples. 

\textbf{Model-specific approaches.} Some works utilize the surrogate model's architecture to improve transferability. Li~\etal~\cite{li2020learning} densely add dropout~\cite{srivastava2014dropout} after each layer to create several ghost networks for better transferability. 
SGM~\cite{wu2020skip} adopts more gradient from the skip connection in ResNets~\cite{he2016resnet} to boost adversarial transferability. LinBP~\cite{guo2020backpropagating} replaces the zeros with ones in the derivative of ReLU to make the model more linear, leading to improved transferability.

\subsection{Adversarial Defense}
Numerous adversarial defenses have been proposed to mitigate the threat of adversarial attacks. Adversarial training~\cite{goodfellow2015FGSM,tramer2018ensemble}, which adopts the adversarial examples during the training process, has been shown as one of the most effective methods~\cite{athalye2018obfuscated} but taking huge computation cost. Recently, Wong~\etal~\cite{wong2020fast} find that single-step adversarial examples can bring satisfiable robustness, making it possible to be applied on large-scale datasets (\eg, ImageNet~\cite{alex2012ImageNet}). Pre-processing the input samples before the model is shown to be another effective way. Liao~\etal~\cite{liao2018defense} design a high-level representation guided denoiser (HGD) to eliminate the adversarial perturbation. Xie~\etal~\cite{xie2018mitigating} find that random resizing and padding on the input image can mitigate the adversarial threat. Naseer~\etal~\cite{naseer2020NRP} train a neural representation purifier (NRP) by a self-supervised adversarial training mechanism to purify the input sample, which exhibits superior effectiveness against transfer-based adversarial examples. On the other hand, certified defense methods aim to provide provable defense in a given radius~\cite{gowal2019scalable,zhang2020towards,cohen2019certified}. For instance, randomized smoothing (RS) trains a robust ImageNet classifier with a tight robustness guarantee~\cite{cohen2019certified}.

\subsection{Data Augmentations}
Data augmentations often transform (\eg, flipping, rotation, cropping, \etc) the image during the training process for better generalization. Mixup~\cite{zhang2018mixup} interpolates two images and their labels to generate virtual samples for training, which also inspires \textit{Admix} to enhance transferability. Cutmix~\cite{yun2019cutmix} pastes an image patch to the original patch and mixes the labels accordingly. AutoAugment~\cite{cubuk2019autoaugment} automatically searches for improved data
augmentation policies (operations and parameters) on the dataset for better generalization, which has been widely adopted in deep learning. Unlike these data augmentation strategies, we aim to construct a set of diverse images by transforming the image block using various transformations, which can be used for gradient calculation to achieve better transferability. 

\section{Methodology}
In this section, we introduce our motivation, provide a detailed description of the proposed
\name, and highlight the difference between \name and AutoAugment.

\begin{table}[t]
    \centering
    \begin{tabular}{cccccc}
        \toprule
         & TIM & DIM & SIM & SSA & \textit{Admix}\\
         \midrule
         Transferability & 57.4 & 77.6 & 79.3 & 80.6 & 83.6\\
        LPIPS & 0.25 & 0.43 & 0.48 & 0.54 & 0.73\\
         \bottomrule
    \end{tabular}
    \caption{The transferability of TIM, DIM, SIM, \textit{Admix}, SSA, and similarity between $1,000$ images and the transformed images evaluated by LPIPS. The transferability is evaluated by the attack success rate of Inception-v3 on the adversarial examples generated on ResNet-18 .}
    \label{tab:similarity}
\end{table}
\subsection{Motivation}

Since Xie~\etal~\cite{xie2019improving} found that transforming the image by random resizing and padding before gradient calculation can generate more transferable adversarial examples, various input transformation based attacks are proposed to further improve transferability. 
As shown in Fig.~\ref{fig:transformation}, the input transformation (\eg, DIM~\cite{xie2019improving} \vs \textit{Admix}~\cite{wang2021admix}) with better transferability tends to bring more obvious visual change to the input image. This inspires us with a new assumption:
  
\begin{assumption}
    Without harming the semantic information, the more diverse the transformed image is, the better transferability the adversarial examples have.
\end{assumption}

\begin{table*}[tb]
    \centering
    \begin{tabular}{ccccccccccc}
         \toprule
         Raw & VShift & HShift & VFlip & HFlip & Rotate & Scale & Add Noise & Resize & DCT & Dropout \\
         \midrule
         \includegraphics[scale=0.095]{figs/pdf/ori.pdf} & \includegraphics[scale=0.195]{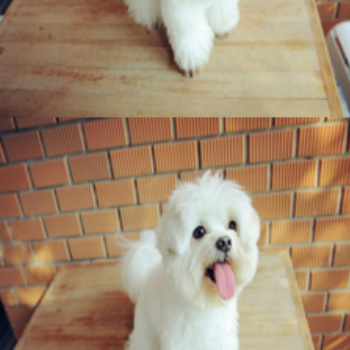} & \includegraphics[scale=0.195]{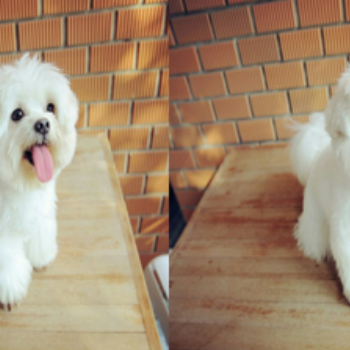} & \includegraphics[scale=0.195]{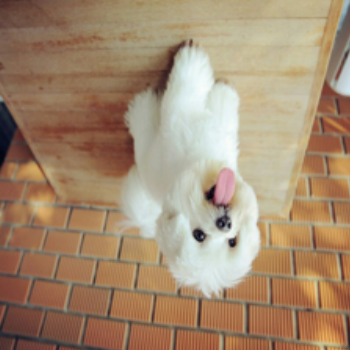} & \includegraphics[scale=0.195]{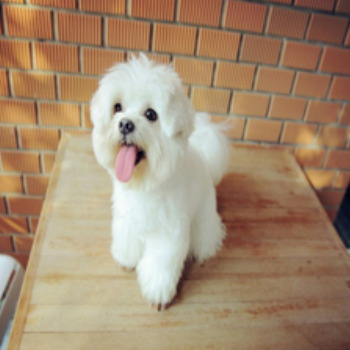} & \includegraphics[scale=0.195]{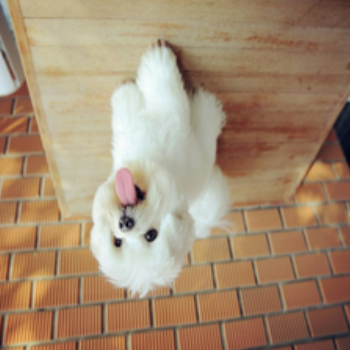} & \includegraphics[scale=0.195]{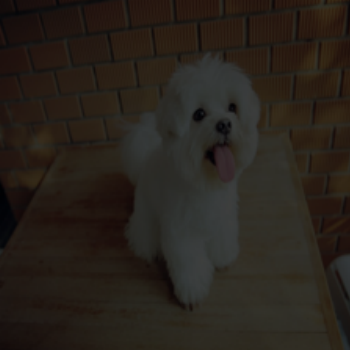} & \includegraphics[scale=0.195]{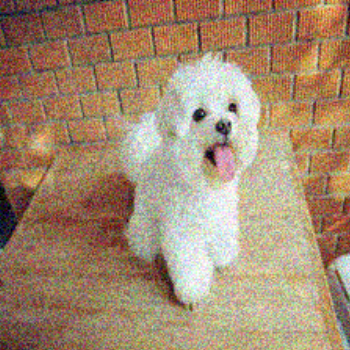} & \includegraphics[scale=0.195]{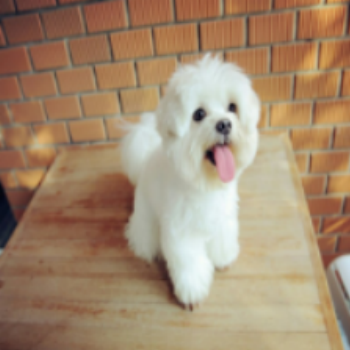} & \includegraphics[scale=0.195]{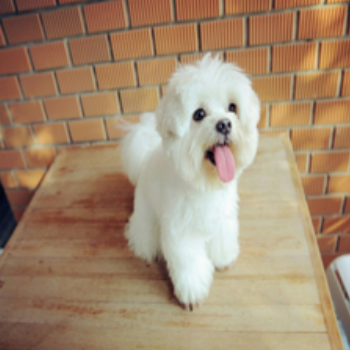} & \includegraphics[scale=0.195]{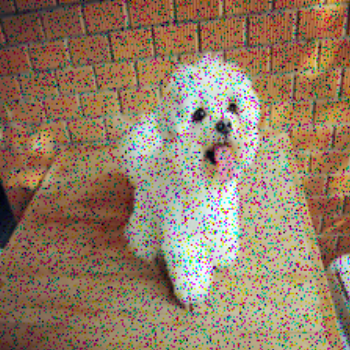} \\
         \bottomrule
    \end{tabular}
    \caption{The raw and transformed images using various transformations adopted by \name (See details in Appendix~\ref{app:imp_details}).}
    \label{tab:transformations}
\end{table*}

To validate this hypothesis, we utilize the learned perceptual image patch similarity (LPIPS)~\cite{zhang2018unreasonable} to evaluate the semantic similarity between the benign samples and the transformed images:
\begin{equation}
    \operatorname{LPIPS}(\bm{x},\hat{\bm{x}})=\frac{1}{H\times W}\sum\limits_{l}\sum\limits_{h,w}\Vert \bm{z}^{l}_{h,w}-\hat{\bm{z}}^{l}_{h,w} \Vert_2
\end{equation}

where $\bm{z}^{l}$ and $\hat{\bm{z}}^{l}$ are the extracted feature from $l$-th layer of SqueezeNet~\cite{hu2018squeeze} with $\bm{x}$ and $\hat{\bm{x}}$ as input, respectively. A smaller LPIPS value indicates better similarity between the two images. The semantic similarity between the raw images and transformed images by DIM, TIM, SIM and \textit{Admix} are summarized in Tab.~\ref{tab:similarity}. As we can see, the similarity decreases when the transferability of attack increases because the transformed images are more diverse. Moreover, as shown in Fig.~\ref{fig:transformation}, such diverse images are significantly different from the raw images, which introduces instability when calculating the gradient. Hence, the more powerful input transformation based attack needs to calculate the gradient on multiple transformed images to eliminate the instability. For instance, DIM utilizes single image, SIM adopts $5$ images, while the SOTA \textit{Admix}/SSA takes $15$/$20$ images.

The relation between the diversity of transformed images and transferability inspires us to generate more diverse images for gradient calculation so that we can craft more transferable adversarial examples. In this work, we apply various input transformations on different blocks of a single input image to obtain more diverse images, detailed in Sec.~\ref{sec:sia}.

\begin{figure}
    \centering
    \begin{minipage}[c]{0.14\textwidth} 
        \begin{subfigure}{\textwidth}
          \centering 
          \includegraphics[width=\linewidth]{figs/pdf/ori.pdf}
          \vspace{-1.3em}
          \caption*{Raw Image}
        \end{subfigure} 
    \end{minipage}
    \hspace{0.1cm}
    \begin{minipage}[c]{0.14\textwidth} 
        \begin{subfigure}{\textwidth}
          \centering 
          \includegraphics[width=\linewidth]{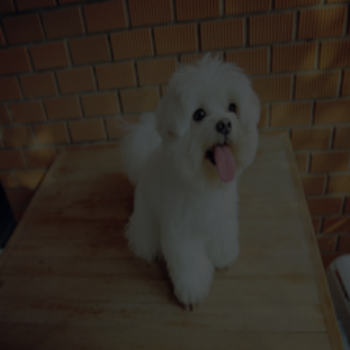}
          \vspace{-1.3em}
          \caption*{Scale}
        \end{subfigure} 
    \end{minipage}
     \hspace{0.1cm}
    \begin{minipage}[c]{0.14\textwidth} 
        \begin{subfigure}{\textwidth}
          \centering 
          \includegraphics[width=\linewidth]{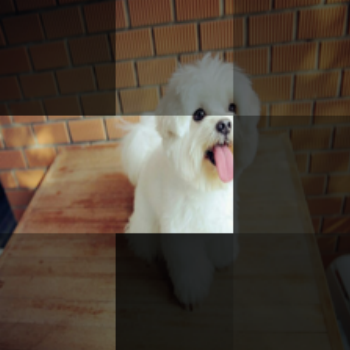}
          \vspace{-1.3em}
          \caption*{Block Scale}
        \end{subfigure} 
    \end{minipage}
    \caption{The randomly sampled raw image and its transformed images by scaling on the full image and $3\times3$ blocks.}
    \label{fig:block_scale}
\end{figure}

\subsection{Structure Invariant Attack}
\label{sec:sia}

Considering the limited number of the existing image transformations, we first investigate that \textit{how to generate a more diverse image using a single transformation?} Without loss of generality, we take scale as an example, which is the base operation of SIM~\cite{lin2020nesterov} (also a particular case of \textit{Admix}~\cite{wang2021admix}). As shown in Fig.~\ref{fig:block_scale}, scaling the image changes its intensity uniformly on all the pixels, resulting in limited diverse images, which is also validated in Tab.~\ref{tab:similarity}. To improve the diversity of scaled images, we scale the image with different factors for different parts. Specifically, we randomly split the image into several image blocks and independently scale each image block with different scaled factors. As illustrated in Fig.~\ref{fig:block_scale}, scaling the image blocks can bring a much more diverse transformed image while humans can still catch the visual information from the image. To explain why such image does not confuse humans, we define the structure of image as follows:
\begin{definition}{(Structure of Image).}
    Given an image $\bm{x}$, which is randomly split into $s\times s$ blocks, the relative relation between each anchor point is the structure of image, where the anchor point is the center of the image block.
\end{definition}
We argue that the structure of image depicts important semantic information for human recognition. For instance, the dog's body should be between its head and tail, while its legs should be under its body. Scaling the image blocks with various factors does not change the structure of image so that the generated image can be correctly recognized by humans as well as deep models.

In summary, transforming the image blocks does not harm the recognition but crafts more diverse images, which is of great benefit to improve transferability. To further boost the diversity of transformed images, we apply various image transformations to different image blocks, where the adopted transformations are summarized in Tab.~\ref{tab:transformations}. To avoid information loss, we add the constraints on some transformations. For instance, the rotation can only rotate the image $180\degree$ to avoid dropping some pixels. We denote such transformation as structure invariant transformation (\tname).

\begin{algorithm}[t] 
\SetKwInput{KwInput}{Input}               
\SetKwInput{KwOutput}{Output}             
    \caption{Structure Invariant Attack}
	\label{alg:sia}
    \LinesNumbered 
    \KwInput{Classifier $\bm{f}(\cdot)$ with the loss function $J$; The benign sample $\bm{x}$ with ground-truth label $y$; The maximum perturbation $\epsilon$, number of iterations $T$ and decay factor $\mu$; Splitting number $s$; Number of transformed images $N$}
	\KwOutput{An adversarial example.}  
    $\alpha=\epsilon/T, \ \bm{g}_0=0, \ \bm{x}_0^{adv} = \bm{x}$\\
    \For{$t = 0 \to T-1$}{
        Constructing a set $\mathcal{X}$ of $N$ transformed images using \tname\\
        Calculating the average gradient on $\mathcal{X}$:
        \begin{equation}
            \bar{\bm{g}}_{t+1} = \frac{1}{N} \sum_{\bm{x}_i \in \mathcal{X}} \nabla_{\bm{x}} J(\bm{x}_i, y)
        \end{equation}\\
        Updating the momentum:
        \begin{equation}
            \bm{g}_{t+1} = \mu \bm{g}_{t} + \frac{\bar{\bm{g}}_{t+1}}{\|\bar{\bm{g}}_{t+1}\|_1}
        \end{equation}\\
        Updating the adversarial example:
        \begin{equation}
            \bm{x}_{t+1}^{adv} = \operatorname{Clip}(\bm{x}_{t}^{adv} + \alpha \cdot \operatorname{sign}(\bm{g}_{t+1}), 0, 1)
        \end{equation}   
    }
    \KwRet{$\bm{x}_{T}^{adv}$}
\end{algorithm}

\begin{figure*}
    \centering
    \includegraphics[width=\textwidth]{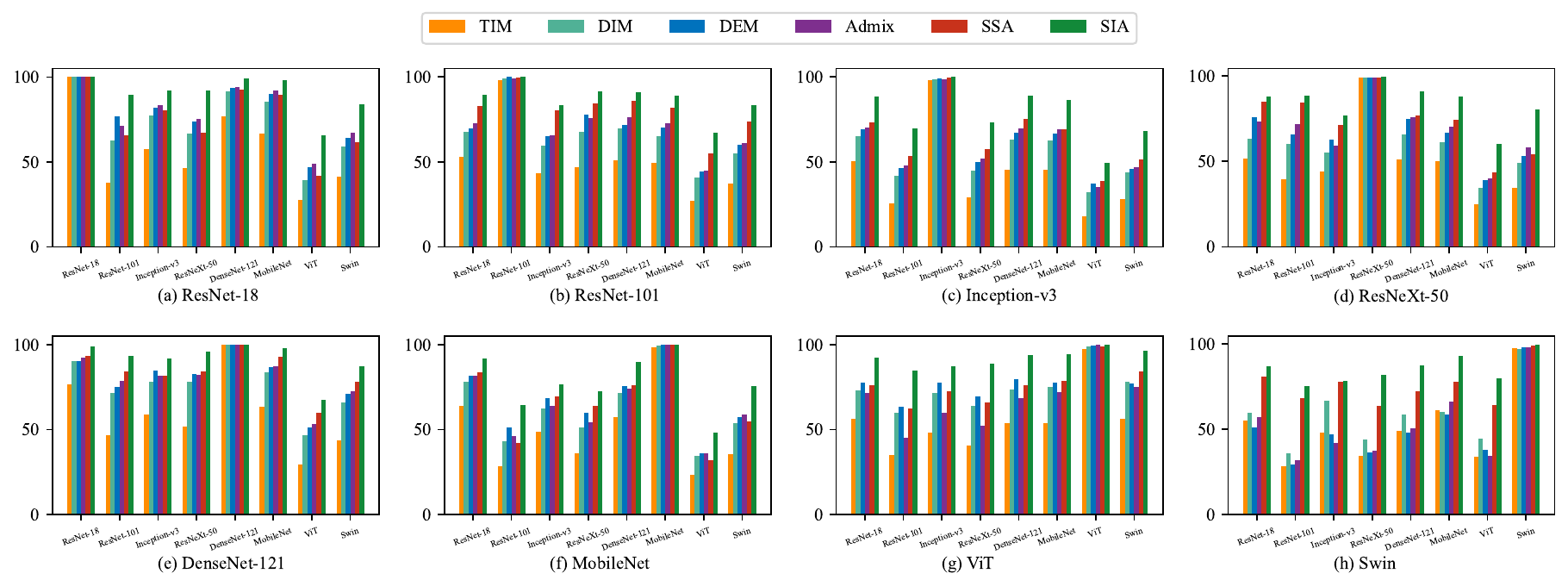}
    \caption{Attack success rates (\%) of eight deep models on the adversarial examples crafted on each model by TIM, DIM, DEM, \textit{Admix}, SSA,  and \name.}
    \label{fig:single_model_attack}
\end{figure*}

Since \tname can generate more diverse images while maintaining the critical semantic information, we adopt \tname as the input transformation to conduct the adversarial attack, denoted as structure invariant attack (\name). Instead of directly calculating the gradient on the input image, \name calculates the gradient on several images transformed by \tname. Note that \name is generally applicable to any gradient-based attacks. Due to the limited space, we integrate \name into MI-FGSM~\cite{dong2018boosting} and summarize the algorithm in Algorithm~\ref{alg:sia}.

\subsection{\tname \vs AutoAugment}
Both \tname and AutoAugment~\cite{cubuk2019autoaugment} leverage several image transformations to pre-process the images. We summarize the difference as follows:
\begin{itemize}[leftmargin=*,noitemsep,topsep=2pt]
\item \tname improves the adversarial transferability when attacking the model while AutoAugment boosts the model generalization during the training process.
\item \tname randomly samples the transformation while AutoAugment needs to search for a good policy for each transformation on each dataset.
\item \tname applies various transformations on different image blocks locally but preserves the global structure. On the contrary, AutoAugment applies two transformations sequentially on the image.
\item The fine-grained transformations by \tname generate more diverse images than AutoAugment.
\end{itemize}
\section{Experiment}
In this section, we conduct extensive evaluations on ImageNet dataset to validate the effectiveness of \name. 

\subsection{Experimental Setting}
\textbf{Dataset.} To align with previous works~\cite{dong2018boosting,lin2020nesterov,wang2021enhancing,wang2021admix}, we randomly sample $1,000$ images pertaining to $1,000$ categories from the ILSVRC 2012 validation set~\cite{Russa2015imagenet}, which are correctly classified by the adopted models.

\textbf{Baselines.} We compare our proposed \name with five competitive input transformation based attacks, namely DIM~\cite{xie2019improving}, TIM~\cite{dong2019evading}, DEM~\cite{zou2020improving}, \textit{Admix}~\cite{wang2021admix}, and SSA~\cite{long2022frequency}, which are integrated into MI-FGSM~\cite{dong2018boosting}. We also integrate the baseline methods with two model-specific approaches, namely LinBP~\cite{guo2020backpropagating} and SGM~\cite{wu2020skip}. 

\textbf{Victim Models.} We evaluate the attack performance on two popular model architectures, namely Convolutional Network Works, \ie ResNet-18~\cite{he2016resnet}, ResNet-101~\cite{he2016resnet}, ResNext-50~\cite{xie2017aggregated}, DenseNet-121~\cite{huang2017densely}, MobileNet~\cite{howard2017mobilenets}, and Transformers, \ie Vision Transformer (ViT)~\cite{dosovitskiy2020image} and Swin Transfromer (Swin)~\cite{liu2021swin}. Furthermore, we study several SOTA defense methods, including one adversarial training method, \ie ensemble adversarially trained model (Inc-v3$_{ens}$)~\cite{tramer2018ensemble}, the top-3 submissions in NIPS 2017 defense competition, \ie high-level representation guided denoiser (HGD)~\cite{liao2018defense}, random resizing
and padding (R\&P)~\cite{xie2018mitigating} and NIPS-r3\footnote{\url{https://github.com/anlthms/nips-2017/tree/master/mmd}}, three input pre-processing based defenses, namely FD~\cite{liu2019feature}, JPEG~\cite{guo2017countering} and Bit-Red~\cite{xu2017feature}, a certified defense, \ie randomized smoothing (RS)~\cite{cohen2019certified} and a deep denoiser, \ie neural representation purifier (NRP)~\cite{naseer2020NRP}.

\textbf{Evaluation Settings.} We follow MI-FGSM~\cite{dong2018boosting} with the perturbation budget $\epsilon=16$, number of iteration $T=10$, step size $\alpha=\epsilon/T=1.6$ and decay factor $\mu=1$. DIM~\cite{xie2019improving} adopts the transformation probability of $0.5$ and TIM~\cite{dong2019evading} utilizes the Guassian kernel with the size of $7\times 7$. DEM~\cite{zou2020improving} uses the resize ratios: $[1.14, 1.27, 1.4, 1.53, 1.66]$. \textit{Admix} admixes $3$ images from other categories with the strength of $0.2$ and $5$ scaled images for each admixed image. SSA~\cite{long2022frequency} sets the turning factor as $0.5$ and the standard deviation as $\epsilon$. \name sets the splitting number $s=3$ and number of transformed images for gradient calculation $N=20$.

\begin{figure}
    \centering
    \includegraphics[width=\linewidth]{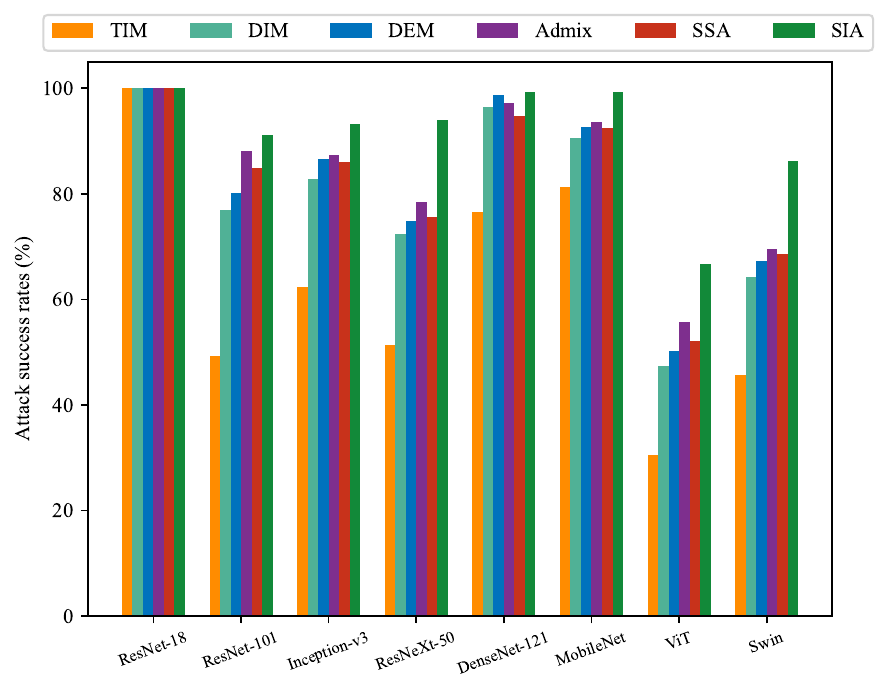}
    \caption{Attack success rates (\%) of eight models on the adversarial examples generated on ResNet-18 when integrating TIM, DIM, DEM, \textit{Admix}, SSA,  and \name into SGM, respectively.}
    \label{fig:sgm}
\end{figure}

\subsection{Attacking a Single Model}
\label{sec:single_model}
To validate the effectiveness of the proposed \name, we first compare \name with five SOTA input transformation based attacks, namely DIM, TIM, DEM,  \textit{Admix} and SSA. We generate the adversarial examples on a single model and test them on the other models. The attack success rates, \ie the misclassification rates of the victim model on the crafted adversarial examples, are summarized in Fig.~\ref{fig:single_model_attack}.

From the figure, we can observe that all the attackers can achieve the attack success rate of $100.0\%$ or near $100.0\%$, showing that the input transformation based attacks do not degrade the white-box attack performance. As for the black-box performance, TIM exhibits the poorest transferability on these normally trained models. \textit{Admix} consistently achieves better transferability than DIM and DEM on CNN-based models. Surprisingly, DIM achieves even better transferability than \textit{Admix} when generating adversarial examples on transformer-based models. SSA exhibits the best transferability among the baselines in most cases. Also, even for the transformer-based models, the adversarial examples generated on Swin show the poorest transferability on ViT than other CNN-based models. Hence, we argue that it is necessary to evaluate the effectiveness of transfer-based attacks on both CNN-based and transformer-based models. We need an in-depth analysis of the transferability among the emerging transformer-based models. Compared with the baselines, \name consistently performs much better than the best baselines on all eight models with different architectures. In particular, \name outperforms the winner-up method with a clear margin of $14.3\%$ on average and achieves an attack success rate higher than the best baseline of at least $2.8\%$ on all the models. Such consistent and superior performance demonstrates that the proposed \name is general to various model architectures (either CNN or transformer) to boost transferability effectively.

\subsection{Integration to model-specific approaches}
To further verify the scalability of the proposed \name, we integrate existing input transformation-based attacks into two model-specific approaches, namely SGM and LinBP. The adversarial examples are generated  on ResNet-18 and test them on the other models.

\begin{figure}
    \centering
    \includegraphics[width=\linewidth]{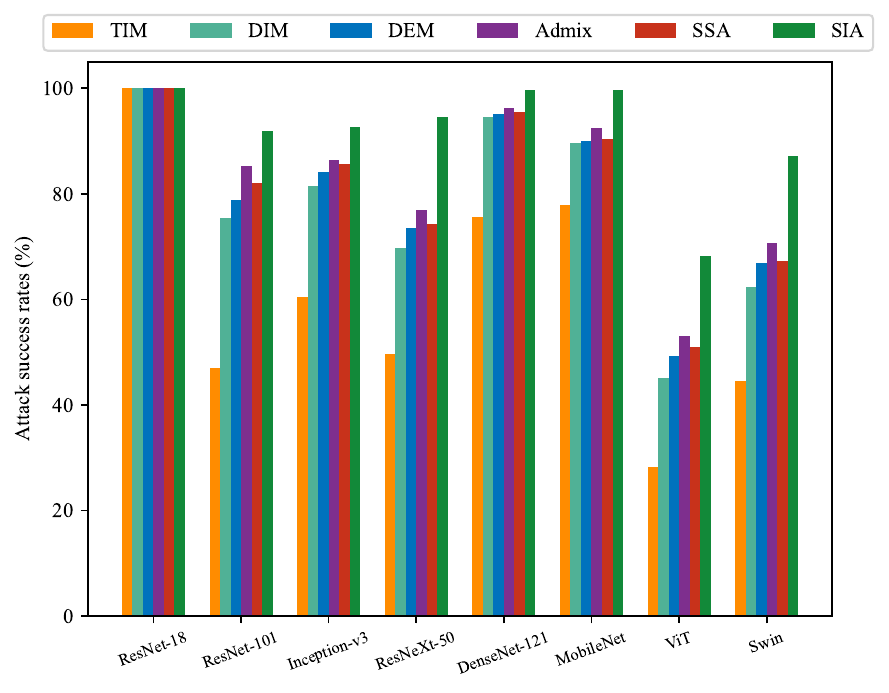}
    \caption{Attack success rates (\%) of eight models on the adversarial examples generated on ResNet-18 when integrating TIM, DIM, DEM, \textit{Admix}, SSA,  and \name into LinBP, respectively.}
    \label{fig:linbp}
\end{figure}

As depicted in Fig.~\ref{fig:sgm} and Fig.~\ref{fig:linbp}, SGM and LinBP can significantly boost the adversarial transferability of vallina input transformation-based attacks. Compared with the baselines, \name still achieves much better transferability, which has been boosted by an average margin of $3.3\%$ and $3.6\%$ when integrated into SGM and LinBP, respectively. These superior results validate the generality of \name to various transfer-based attacks and underscore the potential of \name in augmenting adversarial transferability through the fusion of different strategies. 


\subsection{Attacking Ensemble Models}

\begin{figure}
    \centering
    \includegraphics[width=\linewidth]{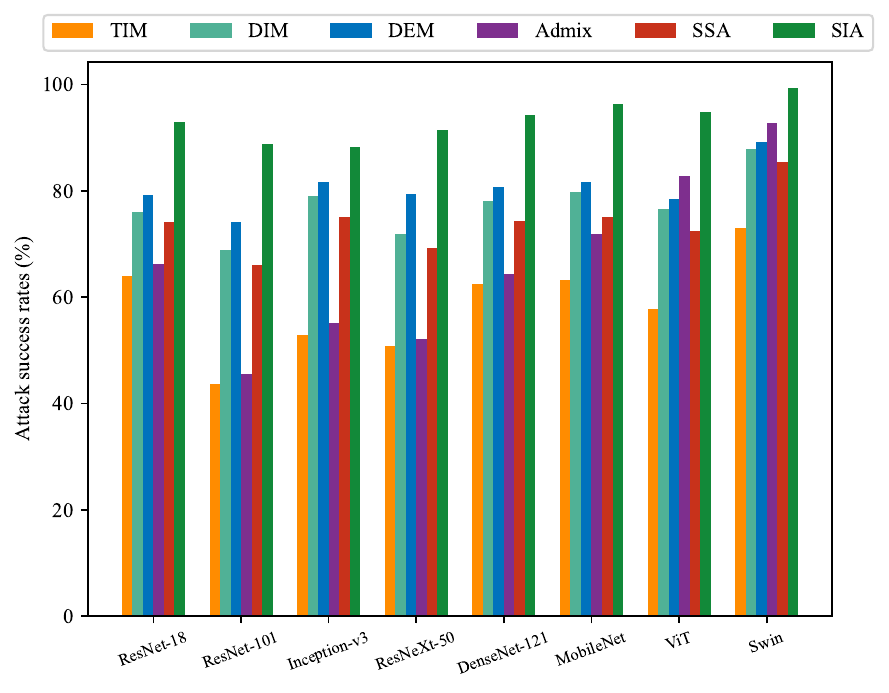}
    \caption{Attack success rates (\%) of eight models on the adversarial examples crafted on ensemble models by TIM, DIM, DEM, \textit{Admix}, SSA,  and \name, respectively. We generate adversarial examples on the CNN-based models and test them on the transformer-based models, and vice versa.}
    \label{fig:ensemble_model}
\end{figure}

Liu~\etal~\cite{liu2016delving} have shown that attacking ensemble models can effectively improve transferability. As shown in Sec.~\ref{sec:single_model}, the adversarial examples often exhibit poorer transferability across the CNN-based models and transformer-based models than being transferred among the CNN-based models. Hence, we generate adversarial examples on ensemble CNN-based and ensemble transformer-based models and test them on the remaining models to evaluate \name when attacking ensemble models.

As shown in Fig.~\ref{fig:ensemble_model}, when attacking ensemble models, the adversarial examples generated by all attacks exhibit better transferability than that crafted on a single model, showing the excellent compatibility of all the input transformation based attacks with such a setting. When generating the adversarial examples on ViT and Swin, DIM and DEM achieve better transferability than \textit{Admix} and SSA, highlighting the difference between crafting adversarial examples on different architectures and the necessity to evaluate the transferability on these models. On all eight models, \name achieves the attack success rate of at least $88.3\%$. It outperforms the winner-up approach with a clear margin of $6.5\%$, showing its superior effectiveness in generating transferable adversarial examples. In particular, \name achieves $94.9\%$ and $99.3\%$ attack success rates on ViT and Swin, respectively, when the adversarial examples are generated on CNN-based models without access to the attention module in the transformer. This further supports our motivation that improving the diversity of transformed images can significantly boost transferability, even on models with completely different architectures.
\begin{figure}
    \centering
    \includegraphics[width=\linewidth] {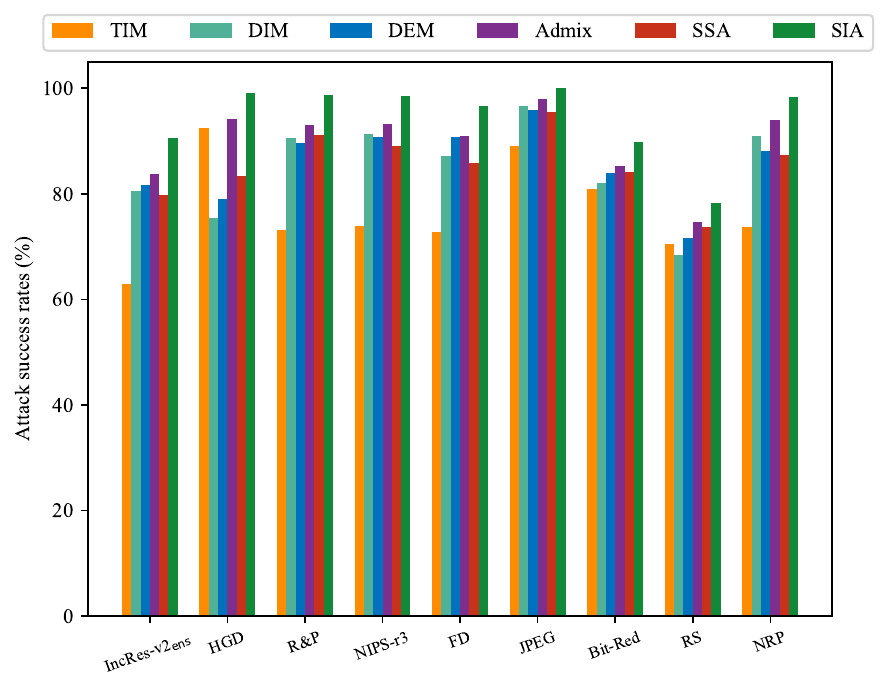}
    \caption{Attack success rates (\%) of various defense methods on the adversarial examples generated by TIM, DIM, DEM, \textit{Admix}, SSA,  and \name under ensemble model setting. The adversarial examples are generated on the eight models simultaneously.}
    \label{fig:defense}
\end{figure}

\subsection{Attacking Defense Methods}
\begin{table*}[tb]
    \centering
    \begin{tabular}{ccccccccccc}
        \toprule
         SIA & -VShift & -HShift & -VFlip & -HFlip & -Rotate & -Sclae & -Add Noise & -Resize & -DCT & -Dropout \\
         \midrule
         92.1 & 89.7 & 90.1 & 90.1 & 88.4 & 88.3 & 90.1 & 90.6 & 90.2 & 90.1 & 90.7\\
         \bottomrule
    \end{tabular}
    \caption{The average attack success rates of adversarial examples crafted by \name and \name without a single transformation. The adversaries are generated on ResNet-18 and tested on the other seven deep models. - indicates removing such transformation.}
    \label{tab:remove_transformation}
\end{table*}

\name has achieved superior attack performance on eight normally trained models with different architectures when attacking single model as well as ensemble models. Recently, several defenses have been proposed to mitigate the threat of adversarial examples on ImageNet dataset. To validate the effectiveness of these defenses, we adopt the adversarial examples generated on these eight models simultaneously to attack the defense models, including Inc-v3$_{ens}$, HGD, R\&P, NIPS-r3, FD, JPEG, Bit-Red, RS and NRP.

The attack results on these defense methods are summarized in Fig.~\ref{fig:defense}. Overall,  DIM is on par with TIM on these defense methods while \textit{Admix} can consistently outperform the other four baselines. \name can consistently achieve better transferability than the baselines. In particular, \name achieves the attack success rate of $78.2\%$ on the certified defense method (\ie RS), and \name can achieve the attack success rate of at least $89.9\%$ on the other eight defense methods, including the powerful denoising method NRP. Such high attack performance indicates the insufficiency of existing defense methods and raises a new security issue to designing more robust deep learning models.

\subsection{Ablation Studies}

To further gain insights on the superior attack performance achieved by \name, we conduct a series of ablation studies to validate that all the transformations and splitting the image into blocks can help improve transferability. We generate the adversarial examples on ResNet-18 and test them on the other seven models for all the experiments.

\begin{figure}
    \centering
    \includegraphics[width=\linewidth]{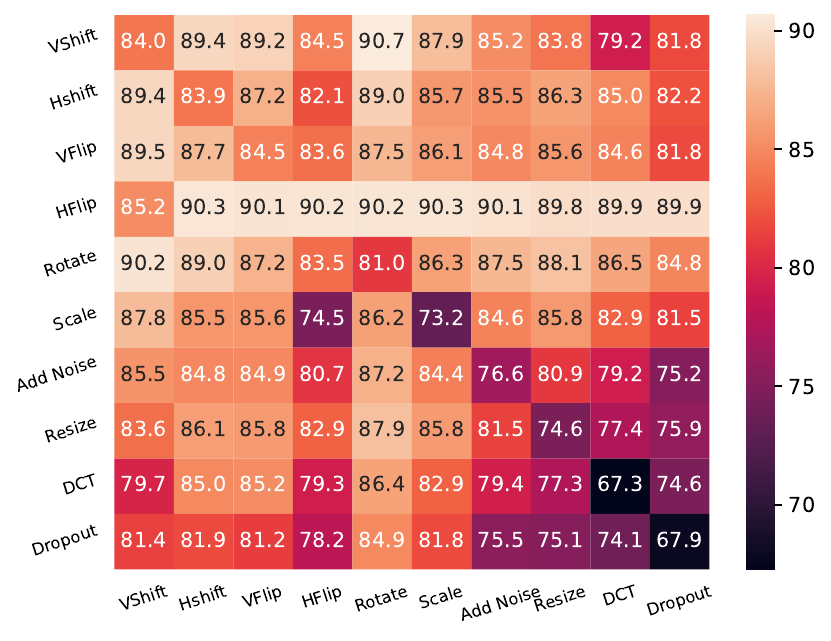}
    \caption{The average attack success rates (\%) on the adversarial examples generated by \name using one (diagonal) or two (other) transformations.}
    \label{fig:ablation_transformation}
\end{figure}

\textbf{Are all the transformations beneficial for boosting the transerability?} There are ten different input transformations that \name might apply to each image block. To investigate whether each transformation is of benefit to generate more transferable adversarial examples, we first adopt one or two transformations to implement \name. The results are summarized in Fig.~\ref{fig:ablation_transformation}. The average attack success rate of MI-FGSM is 59.7\%. On the diagonal line of Fig.~\ref{fig:ablation_transformation}, we only adopt a single transformation for each block and the lowest average attack success rate is $67.3\%$ by adopting DCT, which outperforms MI-FGSM with a margin of 7.6\%. This demonstrates the high effectiveness of \name and its generality to various transformations. We can also observe that the transferability can be further improved when we combine any two transformations, showing the advantage of combining these transformations. To further validate the necessity of each transformation, we generate adversarial examples by \name without each transformation and summarize the results in Tab.~\ref{tab:remove_transformation}. Since removing a single transformation does not significantly decrease the diveristy of the transformation, these attacks achieve similar attack performance. However, no matter which transformation is removed, the transferability will be decreased, supporting that each transformation is of benefit to generate more transferable adversarial examples.

\begin{figure}
    \centering
    \includegraphics[width=\linewidth] {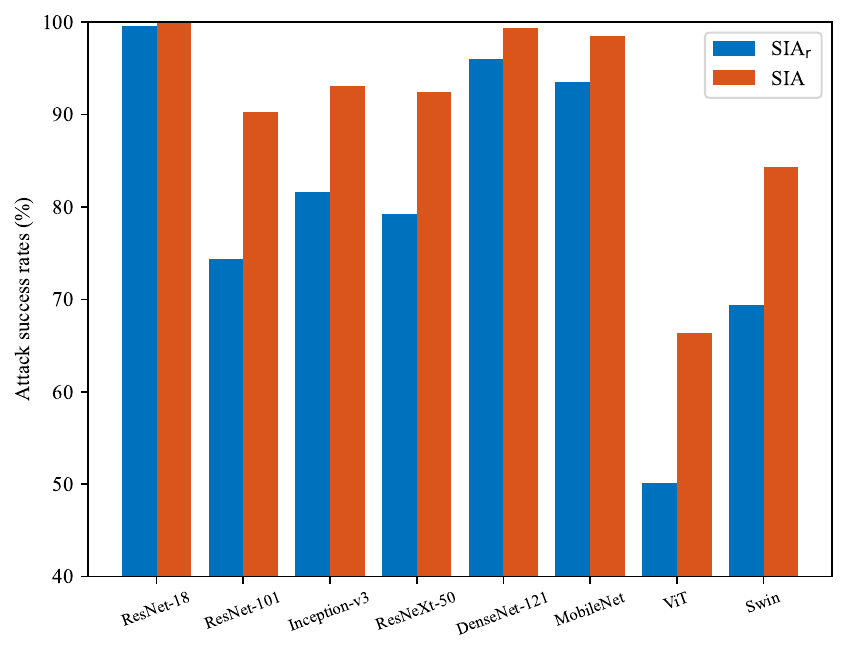}
    \caption{The average attack success rates (\%) of adversarial examples generated by \name and \sname (randomly transform the image without the block partition).} 
    \label{fig:ablation_block}
\end{figure}

\textbf{Are the image blocks beneficial for boosting transferability?} We have shown that the abundant transformations can effectively improve transferability. Here we further explore whether it is useful to apply these transformations to the image block instead of the raw image. For comparison, we randomly apply the transformation to the raw image $s\times s$ times, denoted as \sname. The results are reported in Fig.~\ref{fig:ablation_block}. We can observe that \sname exhibits better transferability than MI-FGSM, which also validates our motivation that improving the diversity of transformed images can generate more transferable adversarial examples. Our \name consistently performs better than \sname, indicating that applying the transformation onto each image block is of great benefit to improve transferability.

\subsection{Parameter Studies}
In this subsection, we conduct parameter studies to explore the impact of two hyper-parameters, namely the number of blocks $s$ and the number of transformed images for gradient calculation $N$. All the adversarial examples are generated on ResNet-18 and tested on the other seven models in the black-box setting.

\begin{figure}
    \centering
    \includegraphics[width=\linewidth]{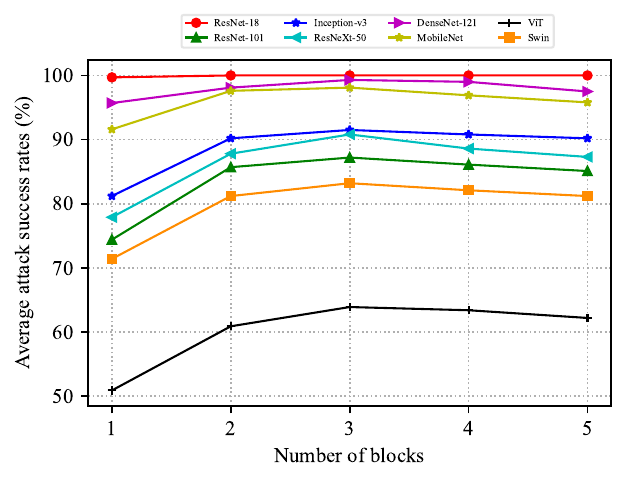}
    \caption{Attack success rates  (\%) of adversarial examples generated by \name with various number of blocks $s$.}
    \label{fig:para_blocks}
\end{figure}

\textbf{On the number of blocks $s$.} The number of blocks determines how diverse the transformed images are, which can influence the attack performance. To find a good value for $s$, we conduct \name with $s$ from $1$ to $5$ and summarize the attack results in Fig.~\ref{fig:para_blocks}. When $s \le 3$, increasing the value of $s$ leads to better diversity, which can improve the attack performance. However, when we continually increase the value of $s$, the diversity is increased, but it also introduces more variance to the gradient, decaying the attack performance slightly. Hence, we adopt $s=3$ to balance the diversity of images and variance of the gradient for better attack performance.

\begin{figure}
    \centering
    \includegraphics[width=\linewidth]{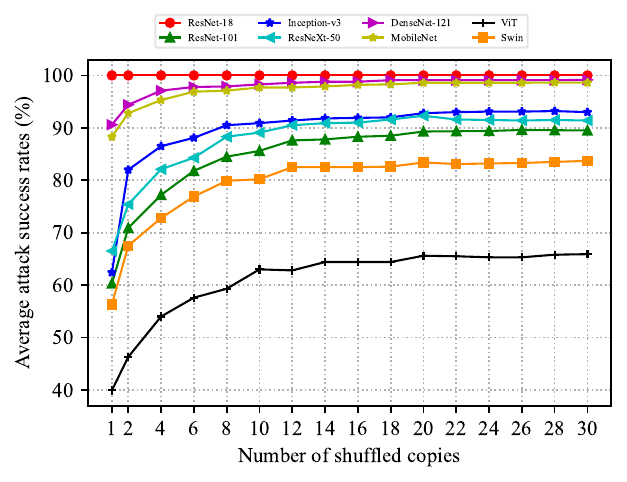}
    \caption{Attack success rates (\%) of adversarial examples generated by \name with various number of images $N$.}
    \label{fig:ablation_number_images}
\end{figure}

\textbf{On the number of images for gradient calculation $N$.} \name calculates the average gradient on $N$ images to eliminate the variance introduced by the transformation. To determine a good value for $N$, we evaluate \name with $N$ from $1$ to $30$ and report the attack success rates in Fig.~\ref{fig:ablation_number_images}. As we can observe, when $N=1$, \name introduces massive variance into the gradient and achieves the lowest performance on all models. When we increase the value of $N$, the variance among the gradients can be eliminated, and \name achieves better transferability before $N=20$. When $N>20$, increasing $N$ can only introduce computation cost without performance improvement. To balance the attck performance and computation cost, we adopt $N=20$ in our experiments.

\section{Conclusion}
In this work, we find that the existing input transformation based attack with better transferability often generates more diverse transformed images. Based on this finding, we design a new image transformation, called structure invariant transformation (\tname), which splits the image into several blocks and randomly transforms each image block independently. With such image transformation, we propose a novel input transformation based attack, dubbed structure invariant attack (\name), which calculates the average gradient on several transformed images by \tname to update the perturbation. Extensive evaluations demonstrate that \name can achieve remarkably better transferability than the existing SOTA attacks. In our opinion, \name provides a new direction by applying the transformation onto the image block to effectively boost transferability, which sheds new light on generating more transferable adversarial examples with fine-grained transformations.

{\small
\bibliographystyle{ieee_fullname}
\bibliography{egbib}
}

\newpage
~
\newpage
\appendix

\section{Implementation details}
\label{app:imp_details}
Our proposed SIA adopts $10$ block-level transformation, namely VShift, Hshift, VFlip, HFlip, Rotate, Scale, Add Noise, Resize, DCT, and Dropout.  Here we provide the implementation details of these transformations, respectively. In particular, for an given image block $\bm{x} \in \mathbb{R}^{3 \times H \times W }$, we can implement the transformations as follows:
\begin{itemize}[leftmargin=*,noitemsep,topsep=2pt]
    \item \textbf{Vertical shift (VShift)}: We roll the image block $\bm{x}$ along the vertical axis by a randomly selected length $h < H$.
    
    \item \textbf{Horizontal shift (Hshift)}: We roll the image block $\bm{x}$ along the horizontal axis by a randomly selected length of $w < W$.
    
    \item \textbf{Vertical flip (VFlip)}: We flip the image block $\bm{x}$ vertically along the horizontal axis, in which the top of the image block becomes the bottom, and the bottom becomes the top. 
    
    \item \textbf{Horizontal flip (HFlip)}: We flip the image block $\bm{x}$ horizontally along the vertical axis, in which the left of the image block becomes the right, and the right becomes the left.
    
    \item \textbf{Rotate}: We turn the image block $\bm{x}$ clockwise by $180\degree$ around its center point, in which the top-left of the image block becomes the bottom-right, and the top-right becomes the bottom-left.
    
    \item \textbf{Scale}: We multiply a random scale factor $\alpha \in (0,1)$ with the pixel in the image block to scale $\bm{x}$ into $\alpha\cdot \bm{x}$.
    
    \item \textbf{Add Noise}: We add a uniform noise $r\in[0,1]^{3\times H\times W}$ to the image block $x$ and clip them into $[0,1]$ to obtain the transformed image block $\mathrm{Clip(x+r, 0, 1)}$.
    \item \textbf{Resize}: We resize $\bm{x}$ into $x'$ with the size of $3\times w\times h$ ($w<W$ and $h<H$) and resize $x'$ back into the size of $3\times W \times H$ using bilinear interpolation.
    \item \textbf{DCT}: We first transform $x$ to the frequency domain using Discrete Cosine Transformation (DCT). Then we mask the top $40\%$ highest frequency with $0$ and recover the image in the time domain using Inverse Discrete Cosine Transformation (IDCT).
    \item \textbf{Dropout}: We randomly mask $10\%$ pixels of $\bm{x}$ with $0$ to obtain the new image block.
\end{itemize}

\end{document}